# Negative Log Likelihood Ratio Loss for Deep Neural Network Classification

Hengshuai Yao[1], Dong-Lai Zhu[1], Bei Jiang[2] and Peng Yu[2]
[1] Huawei Technologies, Canada
[2] Department of Mathematical and Statistical Sciences, University of Alberta, Canada
{hengshuai.yao, donglai.zhu}@huawei.com, {bei1, pyu2}@ualberta.ca

*Abstract*— **In deep neural network, the cross-entropy loss function is commonly used for classification. Minimizing cross-entropy is equivalent to maximizing likelihood under assumptions of uniform feature and class distributions. It belongs to generative training criteria which does not directly discriminate correct class from competing classes. We propose a discriminative loss function with negative log likelihood ratio between correct and competing classes. It significantly outperforms the cross-entropy loss on the CIFAR-10 image classification task.**

## I. Introduction

Deep neural network (DNN) has achieved remarkable success in classification tasks such as image classification [1]. The network output can mimic the posterior probabilities of target classes for the input observation when the non-linear activation function in the output layer is defined as a soft-max function [2]. The learning objective is to minimize the difference between the predicted distribution and the true data-generating distribution. In information theory, the cross entropy between two probability distributions over a common event set of events measures the average number of bits needed to identify an event if coding follows a learned probability distribution rather than the true but unknow distribution [3]. Therefore, cross entropy is a reasonable loss function for the DNN-based classification.

However, in practice the true data-generating probability distribution is unknown and replaced by the empirical probability distribution over a training set where each sample is drawn independently and identically distributed (i.i.d.) from the data space [4]. Under assumptions of uniform distributions of feature and label spaces, minimizing cross-entropy is equivalent to maximum likelihood, i.e., the learning problem aims to maximize likelihood of correct class for each of training samples [2].

Maximum likelihood is a generative training criterion by which the model learns the likelihood of correct class for the observation. The model makes predictions by using Bayes rules to calculate posterior probabilities of target classes for the observation and then select the most likely class. In contrary, discriminative training criteria discriminates the posterior probability of correct class against the competing classes. The model makes predictions with posterior probabilities directly. Therefore, discriminative training criteria are usually preferred to generative training criteria [4].

In this paper we propose a discriminative loss function based on the negative log likelihood ratio (NLLR) between the correct and competing classes for an input feature. It aims to discriminate the posterior probability of correct class against the competing classes. In the DNN learning, the stochastic gradient descent (SGD) algorithm is utilized to minimize the NLLR loss [5]. We evaluate the recognition result on two image classification tasks (MNIST [6] and CIFAR-10 [7]) and compare with the cross-entropy (CE) loss. On the MNIST task, NLLR obtained comparable results as CE because the probability distributions of correct and competing classes are well separated in this relatively easy task. On the CIFAR-10 task, NLLR outperforms CE significantly implying that our discriminative criterion is superior to the generative criterion in classification.

## II. Loss Function for Classification

A classification task is to categorize an observation to one of the target classes. Suppose $x$ is the feature vector representing the observation, and $y$ is the class that the observation belongs to. The goal is to find a function $y = f(x)$ that can predict $y$ from $x$. However, in practical classification tasks there exists the misclassification risk due to missing information, noise interference or probabilistic process. Then the learning problem is to minimize expected risk [4], defined as

$$R(f) = \mathbf{E}_{X,Y \sim p(x,y)} L(f(x), y) \qquad (1)$$

where $L(f(x), y)$ is the loss function measuring the cost of misclassification, and $p(x, y)$ is the probability density function over the feature space X and the class space Y. Because $p(x, y)$ is true but unknown data-generating probability distribution, we draw independently and identically distributed (i.i.d.) samples from the data space and generate a training set

$$S = \{(x_1, y_1), \cdots, (x_m, y_m)\} \qquad (2)$$

which consists of $m$ training samples. Then the expected risk can be approximated by the empirical risk

$$R_S(f) = \frac{1}{m} \sum_{i=1}^{m} L(f(x_i), y_i) \qquad (3)$$

The loss function $L(f(x), y)$ can be flexibly defined to suit the model parameter optimization. For example, in deep neural networks (DNN), the cross entropy is ubiquitous because it measures the difference between the empirical distribution and the predicted distribution, and can be minimized using the stochastic gradient descent (SGD) methods [5]. Suppose there are $C$ target classes in the classification task. The DNN takes the feature vector $x$ as input, and outputs $C$ nodes each representing the score for the corresponding class. When the non-linear activation function in the output layer is defined as a soft-max function, the $C$ outputs can mimic the posterior

.

probabilities of classes $\{\hat{p}(y_c|x); c = 1, \cdots, C\}$ [8]. The cross-entropy loss function is defined as

$$L(f(x), y) = \mathrm{E}_{p(y|x)}[-\log \hat{p}(y|x)]$$
$$= -\sum_{c=1}^{C} p(y_c|x) \log \hat{p}(y_c|x) \quad (4)$$

where $p(y_c|x)$ is the empirical distribution of the training set, and $\hat{p}(y_c|x)$ is the predicted distribution from the DNN model. In classification, each training sample is commonly labeled as the correct class it belongs to, i.e.,

$$p(y_c|x) = \begin{cases} 1 & \text{if } x \in y_c \\ 0 & \text{otherwise} \end{cases} \quad (5)$$

Then Equation (4) can be simplified as

$$L(f(x), y) = -\log \hat{p}(y_c|x) \quad (6)$$

Since the feature distribution $p(x)$ and the class distribution $p(y)$ are irrelevant to the model parameters and assumed to be uniform distributions, according to the Bayesian inference, Equation (6) can be converted to

$$L(f(x), y) = -\log \hat{p}(x|y_c) \quad (7)$$

Equation (7) shows that the loss function is negative log likelihood of sample $x$. Therefore, minimizing the empirical risk Equation (3) is equivalent to maximizing the likelihood. Maximum likelihood is a generative training criterion in which the likelihood score of each training sample is measured. In contrary, the discriminative training criteria attempt to discriminate correct class score from competing class scores for training samples and was preferred as a better direct solution for classification problems [4].

In binary classification task where there are two target classes, a model can be designed with one single output for which the empirical probability $p(y_c|x)$ equals 1 for one class and 0 for the other class. Then Equation (4) can be simplified to a binary cross-entropy loss function [9] as

$$L(f(x), y) = -p(y_c|x) \log \hat{p}(y_c|x)$$
$$-[1 - p(y_c|x)] \log[1 - \hat{p}(y_c|x)] \quad (8)$$

In condition of Equation (5), the binary cross-entropy loss function can be extended to the multi-class classification task as

$$L(f(x), y) = -\log \hat{p}(y_c|x)$$
$$-\sum_{k=1, k \neq c}^{C} \log[1 - \hat{p}(y_k|x)] \quad (9)$$

It can be rewritten as

$$L(f(x), y) = -\log\left[\hat{p}(y_c|x) \prod_{k=1, k \neq c}^{C} \hat{p}(\overline{y_k}|x)\right] \quad (10)$$

where $\hat{p}(\overline{y_k}|x) = 1 - \hat{p}(y_k|x)$ represents the probability of $x$ not classified as $y_k$. Comparing Equation (10) and the cross-entropy loss function Equation (6), it shows that each $\hat{p}(y_c|x)$ is multiplied by a factor $\prod_{k=1, k \neq c}^{C} \hat{p}(\overline{y_k}|x)$ which is product of probabilities of not belonging to each of competing classes. Although Equation (10) consists of probabilities of competing classes, it does not discriminate the correct class probability from the competing ones. We will show that the extended binary cross-entropy loss function cannot improve the performance on multi-class classification tasks.

III. NEGATIVE LOG LIKELIHOOD RATIO LOSS FUNCTION

In this paper we propose an inverse likelihood ratio loss function as a discriminative training criterion. As discussed in the previous section, in practical classification design, minimizing the cross-entropy loss function is equivalent to maximizing likelihood, which is a generative training criterion meaning that only the probability of the correct class is measured for each training sample. It cannot learn to optimize the discrimination between the correct class probabilities and the competing ones. To overcome the shortage of the generative training criterion, we propose to measure the ratio between the predicted correct-class probability and the competing ones in the loss function, defined as

$$L(f(x), y) = -\log \frac{\hat{p}(y_c|x)}{\sum_{k=1, k \neq c}^{C} \hat{p}(y_k|x)} \quad (11)$$

The denominator on the right side is sum of competing-class probabilities representing the probability of not belonging to the correct class, noted as $\hat{p}(\overline{y_c}|x)$. Assuming the feature distribution $p(x)$ and the class distribution $p(y)$ are uniform distributions, like the conversion from Equation (6) to (7) using the Bayesian inference, Equation (11) can be written as

$$L(f(x), y) = -[\log \hat{p}(x|y_c) - \log \hat{p}(x|\overline{y_c})] \quad (12)$$

It shows that the loss function is negative log likelihood ratio between correct and competing classes. It directly learns to discriminate correct class from the competing classes for each training sample. This new loss function achieved significant improvement in our experiments.

IV. EXPERIMENTS

We did experiments on two popular image classification tasks. One is the MNIST task to classify each of 28x28 handwritten digit images into one of 10 digits [6]. The MNIST database has a training set of 60,000 samples, and a test set of 10,000 samples. The second task is on the CIFAR-10 dataset which consists of 60000 32x32 color images in 10 classes, with 6000 images per class [7]. The images are divided into 50000 training images and 10000 test images. For both tasks we train the deep convolutional neural networks (CNN) with the same topologies as the Keras examples [10]. Specifically, for MNIST the CNN topology is defined as two convolutional layers followed by two fully-connected layers with intermediate max-pooling and dropout layers; for CIFAR-10 the CNN topology is defined as four convolutional layers followed by two fully-connected layers with max-pooling and dropout layers. In both networks, the activation function of the output layer adopts the soft-max function to predict the probabilities of target classes on the output nodes.

We compare training and test losses and accuracies over the training epochs for the three loss functions: cross-entropy (CE), binary cross-entropy (BCE), and negative log likelihood ratio (NLLR). For CE and BCE, we train the models with 100 training epochs because MNIST training converges within 100 epochs and CIFAR-10 training encounters overfitting issue within 100 epochs (i.e., accuracy on the test set has started to degrade). For NLLR we train the models with 500 epochs to investigate the generalization performance of the NLLR loss function.

Figure 1 shows the training and test losses and accuracies on the MNIST task. The loss values of NLLR are much lower than the loss values of CE and BCE because NLLR calculates the probability ratio between the correct class and competing classes. MNIST is a relatively easy task where the predicted correct and competing probabilities are mostly close to one and zero respectively, resulting the low loss values. For CE and BCE, the training loss and accuracy curves converge quickly within 100 epochs; the test loss and accuracy converge as well. For NLLR, the training loss and accuracy achieve similar convergence and keep stable in 500 epochs, indicating the robustness of the NLLR loss function. The test loss and accuracy are also convergent and stable within 500 epochs, indicating that the overfitting issue is trivial. Comparing to CE and BCE, NLLR obtained comparable results. Because the correct probability is much bigger than the competing probabilities for most of samples, using NLLR to discriminate the correct probability from the competing probability becomes unimportant for such tasks. For more challenging classification tasks where the correct and competing probabilities could be easily confusing, the NLLR should show different performance from the CE or BCE losses.

CIFAR-10 is a more challenging task than the MNIST. Figure 2 shows the losses and accuracies on the CIFAR-10 task. For CE and BCE, the training loss and accuracy curves converge within 100 epochs; the test loss and accuracy degrade after certain epochs indicating that the training suffers overfitting problems. For NLLR, the training loss and accuracy keep improving over 500 epochs, indicating that the NLLR loss function well fits the SGD optimization; the test loss and accuracy also keep improving over the 500 epochs, indicating that the training does not encounter the overfitting problem. In comparison with CE and BCE, NLLR achieved significant accuracy improvement.

Table 1 summarizes the test accuracies of the three loss functions (CE, BCE and NLLR) on the MNIST and CIFAR-10 tasks. Compared to CE and BCE, NLLR obtained comparable accuracy on the MNIST task and achieved significant improvement on the CIFAR-10 task. It proves that in challenging tasks where correct probabilities are not largely superior to the competing probabilities, the NLLR loss function forms a discriminative criterion yielding better model parameter optimization. Such challenging tasks in practice come from real data with observation noise and variant environment conditions (such as lightning).

TABLE I.     TEST ACCURACIES OF THREE LOSS FUNCTIONS (CE, BCE AND NLLR) ON MNIST AND CIFAR-10 TASKS.

|          | CE     | BCE    | NLLR   |
|----------|--------|--------|--------|
| MNIST    | 99.28% | 99.27% | 99.28% |
| CIFAR-10 | 79.92% | 80.01% | 86.01% |

## V. CONCLUSION

In this paper we first discussed that minimizing cross-entropy is equivalent to maximum likelihood given assumptions of uniform feature and class distributions. It is a generative training criterion where only correct class is measured for each feature sample. We proposed a discriminative loss function NLLR which is a negative log likelihood ratio between correct and competing classes. It aims to maximize the correct class probability and minimize the competing class probabilities simultaneously. Compared to cross-entropy and binary cross-entropy, NLLR obtained comparable results on the MNIST task and significant improvement on the more challenging CIFAR-10 task.

REFERENCES

[1] K. He, X. Zhang, S. Ren, J. Sun, Deep Residual Learning for Image Recognition, CVPR (2016)
[2] I. Goodfellow, Y. Bengio, A. Courville, Deep Learning, MIT Press (2016)
[3] D. Boer, et al., A Tutorial on the Cross-Entropy Method. Annals of Operations Research (2005)
[4] V. Vapnik, The Nature of Statistical Learning Theory, Springer-Verlag (2000).
[5] E. Rumelhart, G. Hinton, J. Williams, "Learning representations by back-propagating errors". Nature (1986).
[6] Y. LeCun, L. Bottou, Y. Bengio, and P. Haffner. "Gradient-based learning applied to document recognition." *Proceedings of the IEEE*, 86(11):2278-2324 (1998).
[7] A. Krizhevsky, Learning Multiple Layers of Features from Tiny Images (2009).
[8] K. Murphy, Machine Learning: A Probabilistic Perspective. MIT (2012).
[9] G. Casella, R. L. Berger, Statistical Inference, Duxbury Thomson Learning (2002).
[10] Keras examples: https://github.com/keras- team/keras/tree/master/

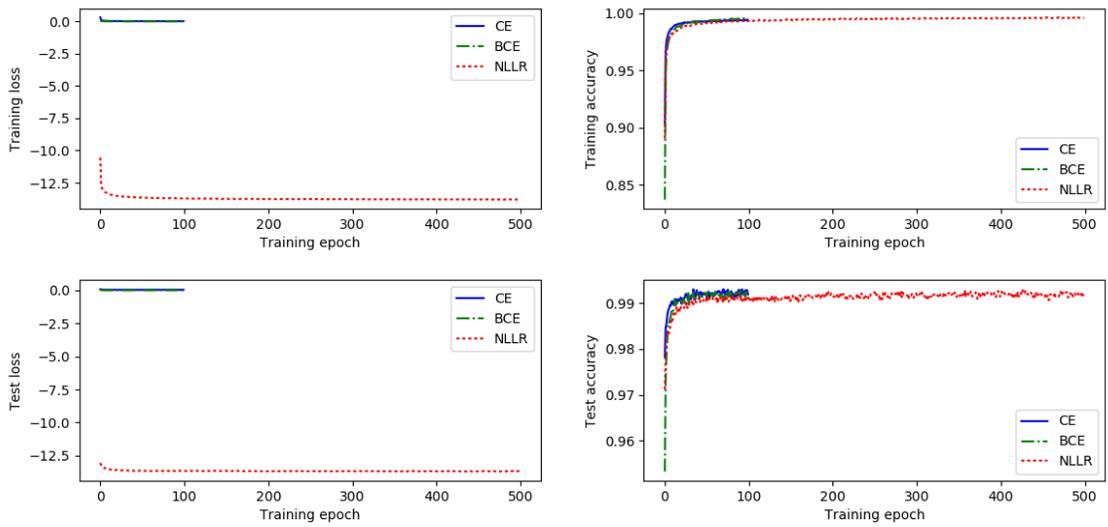

Figure 1. Training, test loss and accuracy of three loss functions (CE, BCE and NLLR) over training epochs on the MNIST task. For NLLR we show 500 epochs while CE and BCE show 100 epochs because curves already become flat.

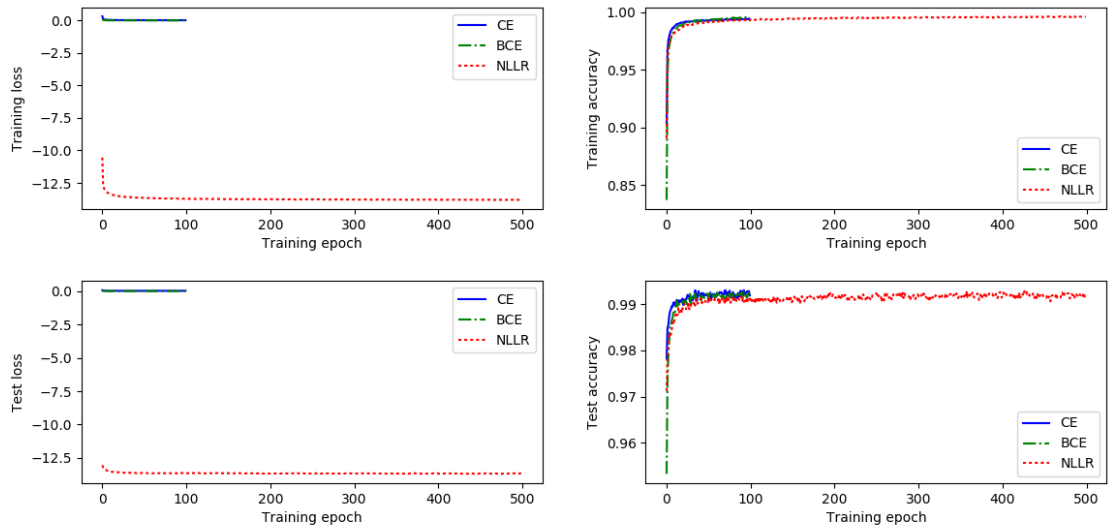

Figure 2. Training, test loss and accuracy of three loss functions (CE, BCE and NLLR) over training epochs on the CIFAR-10 task. For NLLR we show 500 epochs while CE and BCE show 100 epochs because both have encountered overfitting problems within 100 epochs.